\documentclass[]{TEAI}
\usepackage{helvet}

\usepackage{amsmath} 
\usepackage{natbib}
\usepackage{graphicx}
\usepackage{subcaption} 

\usepackage[toc,page,header]{appendix}
\usepackage[utf8]{inputenc} 
\usepackage[T1]{fontenc}    
\usepackage{hyperref}       
\usepackage{url}            
\usepackage{booktabs}       
\usepackage{lmodern}        
\usepackage{amsfonts}       
\usepackage{nicefrac}       
\usepackage{microtype}      
\usepackage{wrapfig}

\usepackage{amssymb}  
\usepackage{fontawesome}  
\usepackage{url}  

\usepackage{titletoc}

\usepackage{tikz}  
\usepackage{comment}  
\usepackage{tabularx}  
\usepackage{booktabs}  

\usepackage{minitoc}

\usepackage{booktabs}
\usepackage{array}
\usepackage{etoolbox}

\definecolor{lightblue}{RGB}{200, 230, 255}  
\definecolor{headerblue}{RGB}{150, 200, 255} 

\usepackage{pgfplots}
\usepackage[utf8]{inputenc} 
\usepackage[T1]{fontenc}    
\usepackage{hyperref}       
\usepackage{url}            
\usepackage{booktabs}       
\usepackage{amsfonts}       
\usepackage{nicefrac}       
\usepackage{microtype}      
\usepackage{xcolor}         
\usepackage{graphicx}
\usepackage{float}
\usepackage{comment}
\usepackage{multirow} 
\usepackage{amsmath} 
\usepackage{makecell} 
\usepackage{siunitx}  
\usepackage{tikz}
\usepackage{pgf-pie} 
\usepackage{subcaption}
\usepackage{wrapfig}
\usepackage[export]{adjustbox}
\usepackage{xspace}

\usepackage{ragged2e}      
\usepackage{tabularx}       
\usepackage{array}          
\usepackage{caption}        
\usepackage{enumitem}
\usepackage{pifont}
\usepackage[hang,flushmargin]{footmisc} 

\usepackage{tcolorbox}

\usepackage{tcolorbox}    
\tcbuselibrary{breakable}  
\tcbuselibrary{skins}      

\usepackage{tabularx}
\usepackage{listings}

\usepackage[ruled,linesnumbered]{algorithm2e}
\usepackage{amsmath}
\usepackage{booktabs}
\usepackage{etoolbox}
\usepackage{graphicx}
\usepackage{makecell}
\usepackage{multirow}
\usepackage{multicol}
\usepackage{physics}
\usepackage{pifont}
\usepackage{tabularx}
\usepackage{xcolor}
\usepackage{xspace}

\newcommand{\CM}{\ding{51}}

\newcommand{\modelName}{HAD\xspace}
\newcommand{\mypara}[1]{\vspace{0.05in}\noindent\textbf{#1}\hspace{0.1in}}

\renewcommand{\arraystretch}{1.1}

\title{HAD: Combining \underline{H}ierarchical Diffusion with Metric-\underline{D}ecoupled RL for End-to-End Driving}

\author{
    Wenhao Yao\textsuperscript{1}, 
    Xinglong Sun\textsuperscript{2}, 
    Zhenxin Li\textsuperscript{1},  
    Shiyi Lan\textsuperscript{2}, 
    Zi Wang\textsuperscript{2}, \\
    Jose M. Alvarez\textsuperscript{2}, 
    Zuxuan Wu\textsuperscript{1,*}
}

\affiliation[1]{\mbox{Fudan University}} 
\affiliation[2]{\mbox{NVIDIA}}

\abstract{
\begin{abstract}

End-to-end planning has emerged as a dominant paradigm for autonomous driving, where recent models often adopt a scoring-selection framework to choose trajectories from a large set of candidates, with diffusion-based decoding showing strong promise. However, directly selecting from the entire candidate space remains difficult to optimize, and Gaussian perturbations used in diffusion often introduce unrealistic trajectories that complicate the denoising process. In addition, for training these models, reinforcement learning (RL) has shown promise, but existing end-to-end RL approaches typically rely on a single coupled reward without structured signals, limiting optimization effectiveness. To address these challenges, we propose \modelName, an end-to-end planning framework with a Hierarchical Diffusion Policy that decomposes planning into a coarse-to-fine process. To improve trajectory generation, we introduce Structure-Preserved Trajectory Expansion, which produces realistic candidates while maintaining kinematic structure. For policy learning, we develop Metric-Decoupled Policy Optimization (MDPO) to enable structured RL optimization across multiple driving objectives. Extensive experiments show that \modelName achieves new state-of-the-art performance on both NAVSIM and HUGSIM, outperforming prior arts by a huge margin: +2.3 EPDMS on NAVSIM and +4.9 Route Completion on HUGSIM.

\end{abstract}
}

\correspondence{\email{whyao23@m.fudan.edu.cn}}

\begin{document}

\maketitle
\renewcommand{\thefootnote}{}
\footnotetext{$^*$ Corresponding author.}
\renewcommand{\thefootnote}{\arabic{footnote}}


\vspace{-1.5em}

\section{Introduction}

Planning is a pivotal component in the autonomous driving pipeline. Recent advances have led to a paradigm shift toward end-to-end planning~\cite{hu2023planning, chen2024vadv2, li2024hydra, liao2024diffusiondrive, zou2025diffusiondrivev2}, where perception, prediction, and planning are integrated into a unified model that directly outputs the driving trajectory from raw sensor inputs. By eliminating intermediate hand-crafted modules, such approaches mitigate error accumulation and enable more holistic reasoning over the driving scene~\cite{hu2023planning}.

To enhance end-to-end planner performance, several works~\cite{chen2024vadv2,li2024hydra,yao2025drivesuprim,li2025ztrs,li2025generalized,liao2024diffusiondrive} adopt a scoring-selection paradigm, where a policy ranks candidate trajectories from discretized action space represented by a fixed vocabulary of anchors (e.g. 8192) and selects the best one. However, performance is fundamentally limited by the coverage and quality of the anchor set, causing failures in out-of-vocabulary scenarios~\cite{liao2024diffusiondrive}. To increase behavioral diversity, recent works~\cite{liao2024diffusiondrive,zou2025diffusiondrivev2,li2026plannerrft,gao2025diffvla++,jiang2025diffvla,li2025recogdrive,liu2025bridgedrive} explore generation with anchored-diffusion, which samples additional trajectories with Gaussian noise perturbations around the trajectory anchors.

\begin{figure*}[!t]
    \centering
    \includegraphics[width=1.0\textwidth]{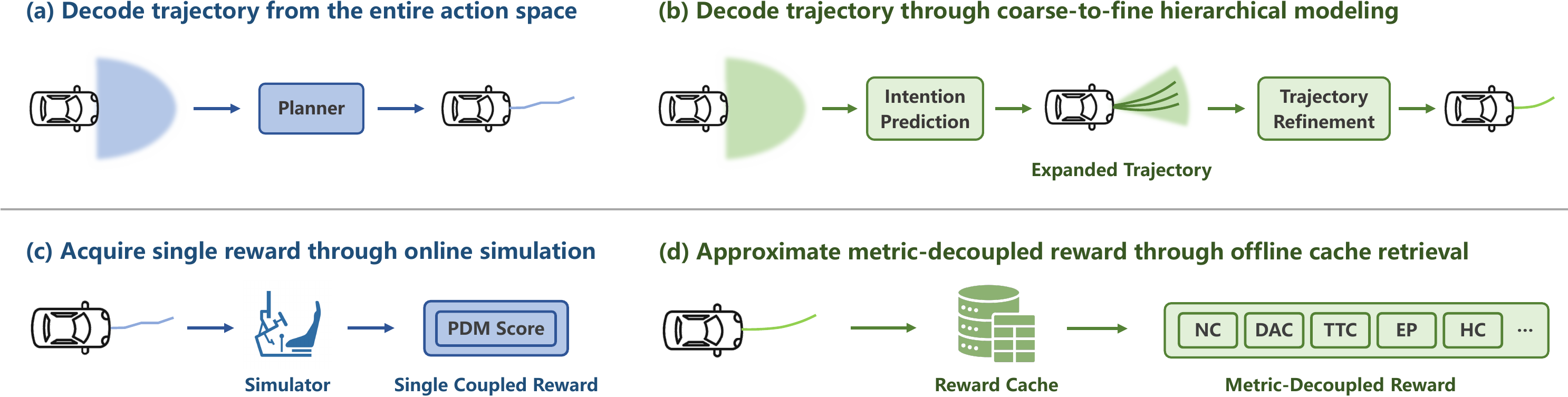}
    \caption{Comparison with existing end-to-end planning methods. Prior approaches search the entire driving space and use a single coupled reward from online simulation. Our method narrows the search via Hierarchical Diffusion Policy and approximates metric-decoupled rewards through offline retrieval.
    }
    \label{fig:teaser}
    \vspace{-10pt}
\end{figure*}

Despite these advances, several limitations remain. First, as shown in Fig.~\ref{fig:teaser} (a), the candidate action space is extremely large, making it a difficult optimization problem for a policy model to directly learn to rank and select from the entire set in a single pass~\cite{yao2025drivesuprim}. Second, naive Gaussian noise sampling is not well suited for driving trajectories. Even in anchored diffusion setups~\cite{liao2024diffusiondrive}, Gaussian perturbations often generate unrealistic motions, such as oscillatory or zigzag trajectories. These low-quality candidates introduce substantial noise into the optimization process and make the denoising stage more challenging.

Moreover, to train these end-to-end policy models, imitation learning~\cite{hu2023planning,chen2024vadv2,liao2024diffusiondrive,chitta2022transfuser} has been the predominant paradigm. However, recent work has shown that imitation learning is fundamentally limited by the quality of human demonstrations~\cite{li2024hydra,li2025ztrs}. Therefore, some works~\cite{zou2025diffusiondrivev2,li2025ztrs,li2026plannerrft} try to incorporate reinforcement learning (RL), where the policy learns to predict trajectories that maximize a reward signal reflecting driving quality, independent of human demonstrations.

Nevertheless, two challenges remain for RL in end-to-end driving. First, existing methods typically rely on a single coupled reward, such as the aggregated PDM score~\cite{dauner2024navsim}. However, driving requires balancing multiple criteria—e.g., collision avoidance, lane keeping, etc—while a single scalar reward is often too coarse and can lead to reward hacking. Recent RL advances in the LLM domain, such as GDPO~\cite{liu2026gdpo}, explore multiple rewards but still aggregate them into a single objective. Second, reward computation must be efficient in practice. Existing approaches~\cite{zou2025diffusiondrivev2,li2026plannerrft} compute rewards on-the-fly via simulation during training, introducing huge computational overhead and reducing training efficiency.

To address the aforementioned limitations, we propose \textbf{\modelName}. To reduce the difficulty of directly decoding a trajectory from the entire action space, we introduce a Hierarchical Diffusion Policy that follows the coarse-to-fine reasoning process of human driving. The policy consists of two diffusion-based denoising stages, as shown in Fig.~\ref{fig:teaser} (b). The first stage identifies a small set of plausible high-level driving intentions represented by trajectory anchors. The second stage then refines these candidates by performing low-level trajectory denoising, effectively “zooming in” on the plausible intentions to produce the final driving maneuver. To generate high-quality trajectory candidates around these intention anchors for low-level trajectory refinement, we further propose Structure-Preserved Trajectory Expansion, which maintains the intrinsic kinematic structure of driving trajectories. Instead of standard Gaussian noise sampling, which often disrupts trajectory structure, we project trajectory anchors from Cartesian space into polar coordinates and apply carefully designed radial scaling and angular offset perturbations. Finally, to better capture the multi-dimensional nature of driving objectives, we introduce Metric-Decoupled Policy Optimization (MDPO), an RL optimization framework that explicitly decouples training into multiple heads, each optimizing a specific driving metric. This design enables more structured and targeted optimization across different aspects of driving behavior. To further improve training efficiency, we also introduce a Trajectory Reward Retrieval mechanism that enables near-instant reward computation via pre-computation and caching of rewards on trajectory anchors combined with training-time nearest-neighbor matching, as shown in Fig.~\ref{fig:teaser} (d).

We comprehensively evaluate \modelName across multiple benchmarks. On the widely used NAVSIM~\cite{dauner2024navsim} benchmark, \modelName achieves 90.2 PDMS on NAVSIM v1 and 88.6 EPDMS on NAVSIM v2, outperforming the previous state-of-the-art method~\cite{jiao2025evadrive} by 2.3 points. In addition, \modelName demonstrates strong closed-loop planning capability on the HUGSIM~\cite{zhou2024hugsim} benchmark, achieving a 47.5 Route Completion rate and a 30.8 HD-Score, surpassing prior art~\cite{li2025ztrs} by 4.9 points in Route Completion and 1.9 points in HD-Score.

Our contribution can be summarized as follows:
\begin{itemize}
    \item We propose Hierarchical Diffusion Policy that decomposes planning into two stages: high-level intention establishment and low-level trajectory refinement, which reduces the optimization difficulty of trajectory selection from large action space.
    \item We propose Structure-Preserved Trajectory Expansion to sample high-quality trajectory candidates, maintaining kinematic structure for denoising.
    \item We develop Metric-Decoupled Policy Optimization (MDPO) to enable fine-grained RL training balancing multiple driving objectives, with an Offline Trajectory Reward Retrieval Scheme to efficiently acquire reward via precomputed anchor rewards and nearest-neighbor retrieval during training.
    \item \modelName achieves achieves state-of-the-art performance on NAVSIM and HUGSIM benchmarks.
\end{itemize}

\section{Related Works}

\subsection{End-to-end Planning}

End-to-end planning methods unify the autonomous driving pipeline from perception to planning into a single network, directly processing raw sensor inputs to output the final driving trajectory.
Early approaches~\cite{hu2023planning, jiang2023vad, shao2023reasonnet, weng2024para, sun2025sparsedrive} apply imitation learning. UniAD~\cite{hu2023planning} was the first work to adopt the planning-oriented pipeline to center all other driving tasks around planning. 
Some methods~\cite{jiang2023vad, sun2025sparsedrive, jia2025drivetransformer} discover sparse scene representation to extract more valuable features for planning.
However, the optimization objective of imitation learning is singular, making models risky of causal confusion problem. To resolve the limitation, 
selection-based methods~\cite{philion2020lift, phan-minh2020covernet, chen2024vadv2, li2024hydra, wang2025enhancing, sima2025centaur, li2025hydranext, yao2025drivesuprim} introduce a predefined trajectory vocabulary, and further use multiple metrics to score each trajectory, selecting the most suitable candidate as the model output. Generative approaches~\cite{zheng2024genad, jiang2023motiondiffuser, zheng2025diffusionbased, xing2025goalflow, liao2024diffusiondrive} utilize VAE~\cite{kingma2014auto} or Diffusion Policy~\cite{chi2023diffusionpolicy} to model dynamic trajectory distribution to reach multi-modal planning.
Recently, to resolve some hard corner cases in driving, some other methods~\cite{shao2024lmdrive, li2025recogdrive, hegde2025distilling} utilize or distill the general VLM~\cite{touvron2023llama, bai2025qwen3vl} knowledge to build robust and instruction-followed planners.
The above methods directly predict trajectories from the whole unstructured, large driving space. Different from these one-stage approaches, our hierarchical modeling pipeline follows coarse-to-fine human driving logic and further eases the causal fusion problem.



\subsection{Reinforcement Learning for Planning}

Several approaches~\cite{toromanoff2020end, jia2023driveadapter, gao2025rad, li2025endtoend, yang2025raw2drive, jiao2025evadrive, zou2025diffusiondrivev2, li2026plannerrft} leverage reinforcement learning (RL) in planning to better align trajectory outputs with pre-defined driving rules. MaRLn~\cite{toromanoff2020end} discovers the potential of RL on planning, but suffers from low training efficiency. RAD~\cite{gao2025rad} establishes a virtual interactive driving environment based on 3DGS to adopt RL training. Some methods like DriveAdapter~\cite{jia2023driveadapter} and Raw2Drive~\cite{yang2025raw2drive} focus on narrowing the gap between raw sensor data and privileged data required for RL training. Other works~\cite{jiao2025evadrive, zou2025diffusiondrivev2, li2026plannerrft} try to integrate RL training with the superior diffusion-based method. EvaDrive~\cite{jiao2025evadrive} introduces a multi-round adversarial RL paradigm to train trajectory generation and scoring models for given scenarios. DiffusionDriveV2~\cite{zou2025diffusiondrivev2} proposes Intra-Anchor GRPO and Inter-Anchor GRPO to improve RL for trajectory anchors. 
These methods require closed-loop driving simulators or reconstructed environments to get a single ensembled trajectory reward, which is computationally expensive and causes policy fusion. Our Trajectory Reward Retrieval Scheme efficiently gets the reward during training, and the Metric-Decoupled Policy Optimization treats the reward as a multi-dimensional vector, improving the policy granularity.

\subsection{Hierarchical Modeling}

Hierarchical modeling is beneficial to progressively refine predictions from coarse to fine, enabling models to handle complex tasks more effectively while improving robustness and stability. It has been widely adopted in various domains of computer vision, like object detection~\cite{ren2015fasterrcnn, cai2018cascade, zhu2021deformable, zhang2023dino_detr}, segmentation~\cite{shi2024part2object, li2022deep}, and image generation~\cite{gafni2022makeascene, ramesh2022hierarchical, razavi2019generating, tian2024visual}.
A few methods~\cite{yao2025drivesuprim, wang2025cognitive, yin2025diffrefiner} employ hierarchical modeling in end-to-end planning. However, they either employ hierarchical modeling in a fixed trajectory set~\cite{yao2025drivesuprim} or ignore modeling detailed exploration in the local region~\cite{wang2025cognitive, yin2025diffrefiner}, resulting in low-quality output trajectories. Our proposed Hierarchical Diffusion Policy explores the local region well through the Structure-preserved Trajectory Expansion Algorithm, leading to accurate output planning trajectories.


\section{Methods}

In this section, we present the proposed method \modelName. We first introduce the Hierarchical Diffusion Policy with the Structure-Preserved Trajectory Expansion algorithm. Next, we present Metric-Decoupled Policy Optimization (MDPO), a reinforcement learning algorithm designed for end-to-end driving. Finally, we introduce an efficient reward retrieval scheme to improve training efficiency.

\subsection{Hierarchical Diffusion Strategy}
\label{subsec:hierarchical_diffusion_strategy}

\begin{figure*}[!t]
    \centering
    \includegraphics[width=\textwidth]{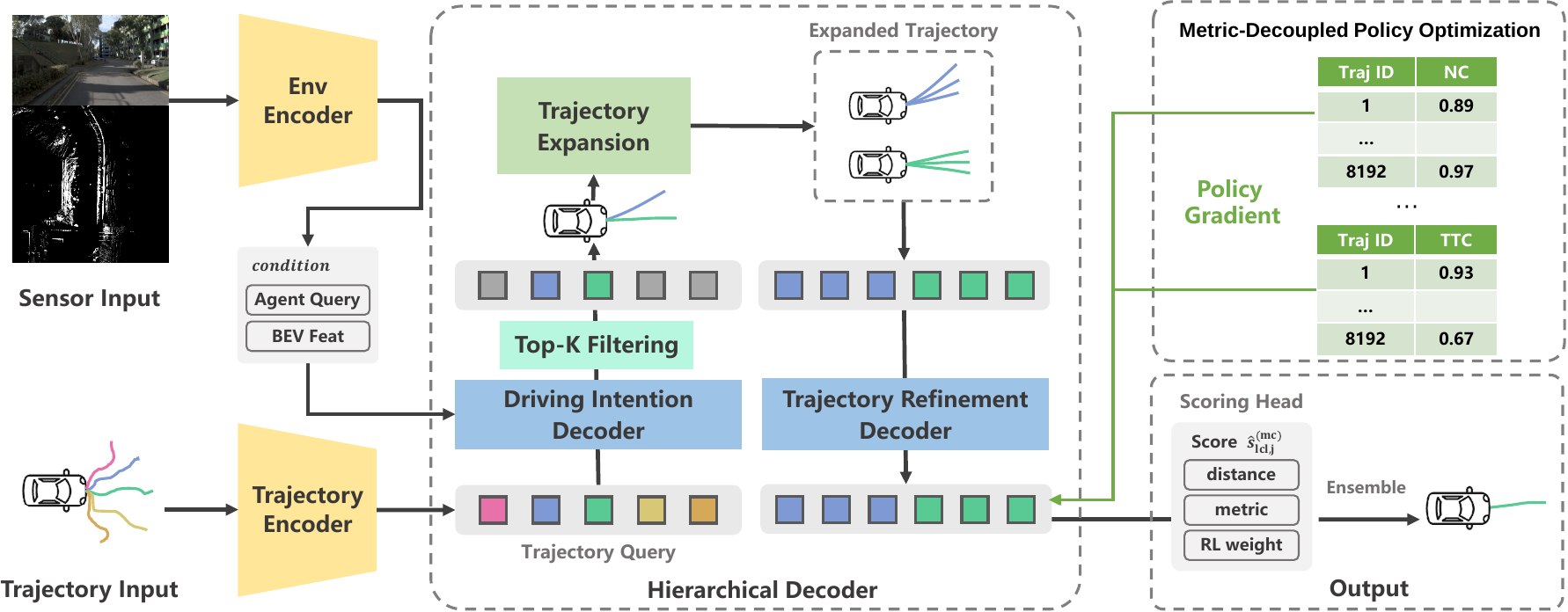}
    \caption{Overview of \modelName. The Hierarchical Diffusion Policy decomposes planning into Driving Intention Establishment and Local Trajectory Refinement. MDPO provides decoupled, structured optimization signals for training.
    }
    \label{fig:arch}
\end{figure*}

\modelName introduces a novel hierarchical diffusion strategy, which treats the trajectory denoising process in a coarse-to-fine manner, decoupling trajectory planning into high-level Driving Intention Establishment and low-level Local Trajectory Refinement.

The framework is illustrated in Fig.~\ref{fig:arch}. First, the Driving Intention Decoder establishes high-level driving intentions by selecting a small set of plausible trajectory anchors. These anchors are then expanded using the Structure-Preserved Trajectory Expansion algorithm to generate diverse local maneuvers around each intention while preserving the intrinsic kinematic structure of trajectories. Second, the Trajectory Refinement Decoder performs fine-grained adjustments within this local region to produce the final trajectory.

\mypara{Driving Intention Establishment}
This stage pre-defines a sparse set of $M_1$ trajectory anchors $\{\tau_j\}_{j=1}^{M_1}$ to roughly cover different driving intentions. During training, these anchors apply a diffusion forward process:
\begin{equation}
\label{equ:add_noise}
    \tilde{\tau}_{j}^{(i)}=\sqrt{\bar{\alpha}^{(i)}} \tau_{j} + \sqrt{1-\bar{\alpha}^{(i)}} \boldsymbol{\epsilon}, \quad \boldsymbol{\epsilon} \sim \mathcal{N}(0, \mathbf{I})
\end{equation}
where $\tau_j=\left\{(x_{j,t}, y_{j,t})\right\}_{t=1}^{T}$ denotes the $j$-th trajectory including $T$ coordinates, $\bar{\alpha}^{(i)}$ represents the cumulative noise schedule at diffusion timestep $i \in [1, T_{\mathrm{trunc}}]$, and $\boldsymbol{\epsilon}$ is the gaussian noise. \modelName leverages a Transformer-based decoder \cite{vaswani2017attention, carion2020detr} to implement trajectory denoising. The encoded perturbed trajectories interact with the environment feature to produce refined trajectories and their corresponding confidence scores:
\begin{gather}
    cond = \mathrm{Enc_{env}}(Img, Lidar, Status) \\
    \left\{f_j\right\}_{j=1}^{M_1} = \mathrm{Enc_{traj}}\left(\left\{\tilde{\tau}_j\right\}_{j=1}^{M_1}\right) \\
    \left\{(\hat\tau_{\mathrm{gbl},j}, \hat{s}_{\mathrm{gbl}, j}^{(c)})\right\}_{j=1}^{M_1} = \mathrm{Dec_{gbl}}\left(\left\{f_j\right\}_{j=1}^{M_1}, cond\right)
\end{gather}
where $\mathrm{Enc_{env}}$ is a multi-modal environment encoder identical to the Transfuser \cite{chitta2022transfuser} backbone, fusing features from images $Img$, LiDAR $Lidar$ and ego-vehicle status $Status$ to extract environmental feature $cond$, $f_j$ is the $j$-th trajectory query encoded by trajectory MLP encoder $\mathrm{Enc_{traj}}$, and $\mathrm{Dec_{gbl}}$ is the Driving Intention Decoder. The decoder yields two critical outputs: the denoised trajectory $\hat\tau_{\mathrm{gbl},j}$ and the classification score $\hat{s}_{\mathrm{gbl}, j}^{(c)}$. The score reveals whether the denoised trajectory is in the same driving sub-region as the human trajectory. We select candidates with top-$K$ classification score as the filtered trajectory set $\mathcal{T}_{\mathrm{gbl}} = \{\hat{\tau}_{\mathrm{gbl},j} \mid j \in \mathcal{I}_{\text{topK}}\}$, where $\mathcal{I}_{\text{topK}} = \operatorname{argsort}_K(\{\hat{s}_{\mathrm{gbl}, j}^{(c)}\}_{j=1}^{M_1})$, to represent driving intention.


\mypara{Structure-Preserved Trajectory Expansion}
To ensure a comprehensive exploration of the local driving region around the plausible intentions, \modelName incorporates a Structure-Preserved Trajectory Expansion algorithm. This approach enriches trajectory diversity while strictly maintaining the intrinsic kinematic trajectory structure, easing the risk of trajectory structure damage caused by naive Gaussian noise sampling~\cite{liao2024diffusiondrive,zou2025diffusiondrivev2,li2026plannerrft,chi2023diffusionpolicy}. Given a trajectory $\tau=\{(x_t, y_t)\}_{t=1}^T$ and an expansion number $n_\mathrm{exp}$, this algorithm outputs an expanded trajectory set $\mathcal{T}_{\mathrm{exp}} = \{ \tau_{j}\}_{j=1}^{n_{\mathrm{exp}} \cdot n_{\mathrm{exp}}}$. Specifically, the algorithm first projects the Cartesian coordinates of $\tau$ into the \textit{polar domain}, represented as $\tau=\{(\rho_t, \theta_t)\}_{t=1}^T$. Moreover, we define a set of radial scaling coefficients $\{\lambda_u\}_{u=1}^{n_\mathrm{exp}}$ and angular offset coefficients $\{\delta_v\}_{v=1}^{n_\mathrm{exp}}$. The algorithm traverses these coefficients to apply linear transformations in the polar space, resulting in the expanded trajectories:
\begin{gather}
    \rho_t^{(u,v)} = \lambda_u \cdot \rho_t, \quad \theta_t^{(u,v)} = \theta_t + \delta_v \\
    x_t^{(u,v)} = \rho_t^{(u,v)} \cos\theta_t^{(u,v)}, \quad y_t^{(u,v)} = \rho_t^{(u,v)} \sin\theta_t^{(u,v)}
\end{gather}
The resulting trajectory is denoted as $\tau^{(u,v)}=\{(x_t^{(u,v)}, y_t^{(u,v)})\}_{t=1}^T$. The more detailed algorithm process is listed in Appendix~\ref{sec:appendix-traj_expansion}.

We apply the trajectory expansion algorithm to the top-$K$ candidates $\mathcal{T}_{\mathrm{gbl}}$, yielding an expanded set $\mathcal{T}_{\mathrm{gbl\text{-}exp}}=\{ \tau_{j}\}_{j=1}^{M_2}$, where $M_2 = K \cdot n_{\mathrm{exp}}^2$ denotes the total number of trajectories in the second stage.

\mypara{Local Trajectory Refinement}
This stage performs fine-grained trajectory refinement over the expanded candidates to produce the final refined local driving maneuver. Given the expanded trajectory set $\mathcal{T}_{\mathrm{gbl\text{-}exp}}=\{ \tau_{j}\}_{j=1}^{M_2}$, the model employs the Trajectory Refinement Decoder $\mathrm{Dec_{lcl}}$ to refine these candidates:
\begin{gather}
    \left\{g_j\right\}_{j=1}^{M_2} = \mathrm{Enc_{traj}}\left(\left\{\tau_j\right\}_{j=1}^{M_2}\right) \\
    \label{equ:dec_lcl}
    \left\{(\hat\tau_{\mathrm{lcl},j}, \hat{s}_{\mathrm{lcl}, j}^{(\mathrm{mc})})\right\}_{j=1}^{M_2} = \mathrm{Dec_{lcl}}\left(\left\{g_j\right\}_{j=1}^{M_2}, cond\right)
\end{gather}
where $cond$ represents the environmental features extracted in the first denoising stage, $\tau_j$ denotes the expanded trajectory, $g_j$ is the encoded expanded trajectory query, and $\hat\tau_{\mathrm{lcl},j}$ is the refined trajectory.

Besides decoding trajectories, \modelName introduces multiple MLP heads in $\mathrm{Dec_{lcl}}$ to predict different metrics for each decoded trajectory, denoted as $\hat{s}_{\mathrm{lcl}, j}^{(\mathrm{mc})}$. Specifically, the metric set $\mathrm{mc}$ covers three categories: (i) $\hat{s}_{\mathrm{lcl}, j}^{(\mathrm{dist})}$, the distance to the expert trajectory, (ii) $\hat{s}_{\mathrm{lcl}, j}^{(m)}$, safety-critical metrics defined in the NAVSIM benchmark \cite{dauner2024navsim}, and (iii) $\hat{s}_{\mathrm{lcl}, j}^{(m\text{-}\mathrm{rl})}$, the metric-decoupled reinforcement learning weights  (detailed in Sec.~\ref{subsec:methods-rl-alg}). The final planning output is derived through a weighted average of the refined candidates:
\begin{gather}
    \label{equ:pdms_score_infer}
    \hat{s}_{\mathrm{lcl}, j}^{(\mathrm{pdms})} = \sum_{mp}{\lambda^{(mp)}\log {\hat{s}_{\mathrm{lcl}, j}^{(mp)}}} +\lambda_{\mathrm{avg}}\log\left({\sum_{ma}{\lambda^{(ma)}s_{\mathrm{lcl}, j}^{(ma)}}}\right) \\
    \label{equ:lcl_score_infer}
    \hat{s}_{\mathrm{lcl}, j} = \gamma_{\mathrm{dist}}\hat{s}_{\mathrm{lcl}, j}^{(\mathrm{dist})} + \gamma_{\mathrm{pdms}}\hat{s}_{\mathrm{lcl}, j}^{(\mathrm{pdms})} + \gamma_{\mathrm{rl}}\hat{s}_{\mathrm{lcl}, j}^{(\mathrm{rl})} \\
    \hat{\tau} = \sum_j w_j \hat{\tau}_{\mathrm{lcl}, j}, \quad\text{where } w_j=\mathtt{Softmax}(\hat{s}_{\mathrm{lcl}, j})
\end{gather}
where $\hat{s}_{\mathrm{lcl}, j}^{(\mathrm{pdms})}$ is the linear combination of logarithm of different metric scores $\hat{s}_{\mathrm{lcl}, j}^{(m)}$, $mp$ and $ma$ are penalty metrics and average metrics. $\hat{s}_{\mathrm{lcl}, j}^{(\mathrm{rl})}$ ensembles the metric-decoupled reinforcement learning weights $\hat{s}_{\mathrm{lcl}, j}^{(m\text{-}\mathrm{rl})}$, following similar score ensemble procedure as $\hat{s}_{\mathrm{lcl}, j}^{(\mathrm{pdms})}$. $\gamma_{\mathrm{dist}}$, $\gamma_{\mathrm{pdms}}$, and $\gamma_{\mathrm{rl}}$ are coefficients, and $\hat{\tau}_{\mathrm{lcl}, j}$ is the trajectory output from the Local Trajectory Refinement stage.

\subsection{Metric-Decoupled Policy Optimization}
\label{subsec:methods-rl-alg}

To better align diffusion-decoded trajectories with safety-related driving rules, we employ reinforcement learning for policy optimization. Prior end-to-end RL approaches, such as DiffusionDriveV2~\cite{zou2025diffusiondrivev2}, typically rely on a single aggregated reward representing overall trajectory quality. However, such a coarse reward signal can hinder effective optimization and may lead to reward hacking~\cite{skalse2022defining}. To address this issue, we propose Metric-Decoupled Policy Optimization (MDPO), which enables more structured optimization across multiple driving metrics.

Specifically, the first stage of the Hierarchical Diffusion Policy generates $K$ candidate trajectories. Through a trajectory expansion algorithm, these candidates are transformed into $K$ sets of local trajectories $\{\mathcal{T}_{\text{gbl-exp}, k}\}_{k=1}^K$, where each set contains $M_{\mathrm{sub}} = n_{\mathrm{exp}}^2$ trajectories. For the $k$-th set $\mathcal{T}_{\text{gbl-exp}, k}=\left\{\hat\tau_{\mathrm{lcl},j}\right\}_{j \in \mathcal{I}_k}$, MDPO first computes the selection probability $p_{j}^{(m)}$ for each trajectory on each safety metric~\cite{dauner2024navsim}, then calculates the reward $J_k$ for the $k$-th trajectory set:
\begin{gather} 
p_j^{(m)} = \frac{\exp \left( \hat{s}_{\mathrm{lcl}, j}^{(m\text{-}\mathrm{rl})} \right)}{\sum\limits_{i \in \mathcal{I}_k} \exp \left( \hat{s}_{\mathrm{lcl}, i}^{(m\text{-}\mathrm{rl})} \right)} \\
\bar{r}_j^{(m)} = \frac{r_j^{(m)} - \operatorname{mean}\left(\{ r_i^{(m)} \mid i \in \mathcal{I}_k \}\right)}{\operatorname{std}\left(\{ r_i^{(m)} \mid i \in \mathcal{I}_k \}\right) + \epsilon} \\
\label{equ:reward_j_k}
J_k =  \sum\limits_{m}{\alpha^{(m)} \sum_{j \in \mathcal{I}_k} p_{j}^{(m)} \cdot\bar{r}_j^{(m)}}
\end{gather}
where $m$ denotes the safety metrics, and $\hat{s}_{\mathrm{lcl}, j}^{(m\text{-}\mathrm{rl})}$ is the RL logit predicted by the model for the $j$-th trajectory regarding the $m$-th metric. $\bar{r}_j^{(m)}$ represents the normalized reward for metric $m$, $\mathcal{I}_k$ denotes trajectory indices in the $k$-th set. Finally, $\alpha^{(m)}$ is the predefined coefficient for each metric, $J_k$ is the reward of the $k$-th trajectory set. The total reward $J$ is defined as the mean reward across all trajectory sets: $J = \frac{1}{K} \sum_{k=1}^K J_k$.


\subsection{Efficient Offline Trajectory Reward Retrieval}
\label{subsec:trajectory_reward_retrieval}

\begin{figure*}[!t]
    \centering
    \includegraphics[width=\textwidth]{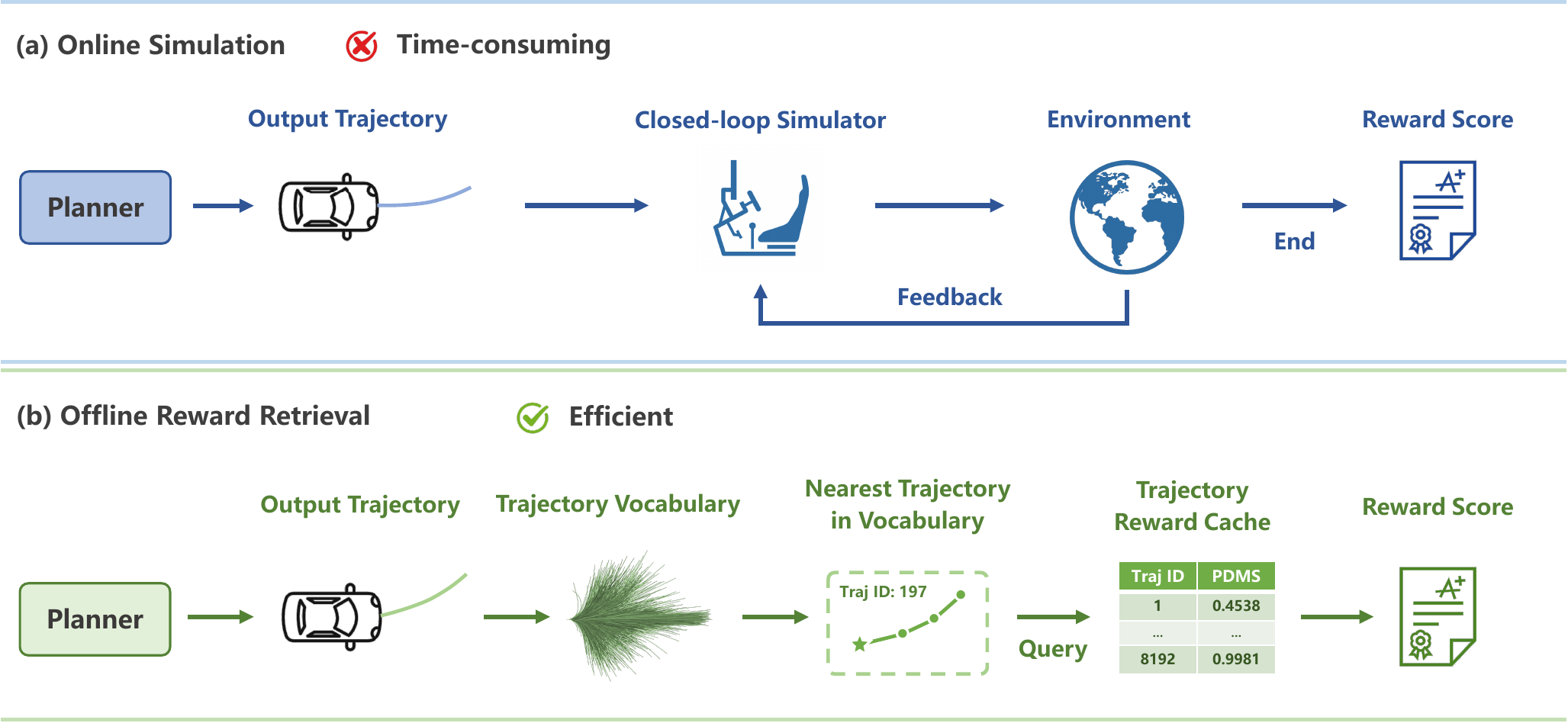}
    \caption{Comparison between simulation-based trajectory evaluation and our proposed Offline Reward Retrieval scheme.
    } 
    \label{fig:area_discretization}
\end{figure*}

Driving reward scores typically require running a simulator to compute safety metrics such as collision avoidance~\cite{zou2025diffusiondrivev2}. However, such a simulation is a CPU-intensive task and time-consuming, making it inefficient and difficult to scale when executed on-the-fly during training, as illustrated in Fig.~\ref{fig:area_discretization}~(a). Therefore, we propose an efficient Offline Trajectory Reward Retrieval Scheme based on driving area discretization, as shown in Fig.~\ref{fig:area_discretization} (b). Inspired by recent trajectory-selection frameworks~\cite{li2024hydra, li2025hydranext}, \modelName incorporates a extensive trajectory vocabulary $\mathcal{V} = \{ \tau_{\mathrm{ref}, j} \}_{j=1}^N$, providing dense $N$ trajectories covering the whole feasible driving area. For each trajectory $\tau_{\mathrm{ref}, j}$ in the vocabulary $\mathcal{V}$, we calculate its safety metric scores $s_{\mathrm{ref}, j}^{(m)}$ in advance, and these scores are stored in a trajectory reward caching table. During online training, for an arbitrary output trajectory $\hat{\tau}$ predicted by the model, our scheme performs a nearest-neighbor query to find the closest reference trajectory $\tau_{\mathrm{ref}, j^*}$ in $\mathcal{V}$:
\begin{equation}
\tau_{\mathrm{ref}, j^*} = \underset{\tau_{\mathrm{ref}, j} \in \mathcal{V}}{\operatorname{arg\,min\,}} \mathrm{Dist}(\hat{\tau}, \tau_{\mathrm{ref}, j})
\end{equation}
The corresponding scores $s_{\mathrm{ref}, j^*}^{(m)}$ are directly utilized as the approximation of the simulation score of $\hat{\tau}$. 

This strategy converts expensive online simulation into an efficient nearest-neighbor trajectory retrieval from the offline cache. As a result, reinforcement learning rewards can be obtained much more efficiently, significantly improving training efficiency.

\subsection{Loss Function}

The loss function of \modelName consists of three components: the perception loss $L_{\mathrm{percept}}$, the driving intent establishment loss $L_{\mathrm{global}}$, and the local trajectory refinement loss $L_{\mathrm{local}}$:
\begin{equation}
    \label{equ:loss_all}
    L = \lambda_{\mathrm{percept}}L_{\mathrm{percept}} + \lambda_{\mathrm{global}}L_{\mathrm{global}} + \lambda_{\mathrm{local}}L_{\mathrm{local}}
\end{equation}

The perception loss $L_{\mathrm{percept}}$ follows the design of Transfuser~\cite{chitta2022transfuser}, including the detection loss for 3D bounding box regression, the classification loss for object categories, and the semantic segmentation loss on bird's-eye-view (BEV) representation. The driving intent establishment loss $L_{\mathrm{global}}$ is employed to supervise the output of the first-stage decoder $\mathrm{Dec_{gbl}}$, comprising the trajectory regression loss and the region classification loss, which determines whether the predicted trajectory and the ground-truth human trajectory belong to the same spatial region. The local trajectory refinement loss $L_{\mathrm{local}}$ aims to supervise the refined trajectories and their multi-dimensional quality metrics produced in the second stage, including the distance to the human expert trajectory, supervision for safety metrics, and the reinforcement learning (RL) objective. More details about the loss function are listed in Appendix~\ref{sec:appendix-loss_func}.

\section{Experiments}

\subsection{Implementation Details}
\label{subsec:imple_details}

\mypara{Datasets}
The experiments are mainly conducted on two datasets: NAVSIM~\cite{dauner2024navsim} and HUGSIM~\cite{zhou2024hugsim}.
NAVSIM is an open-loop planning benchmark. The output trajectory is judged by a simulator to determine whether this trajectory follows driving rules. The evaluation metric of NAVSIM is PDMS, which is aggregated by several driving sub-metrics:
\begin{equation}
    \mathrm{PDMS} = \left( \prod_{m \in S_{\mathrm{pen}}}{\mathrm{s}_m} \right)  \times \left( \frac{ \sum_{w \in S_{\mathrm{avg}}}{\mathrm{w}_w \times \mathrm{s}_w} }{ \sum_{w \in S_{\mathrm{avg}}}{\mathrm{w}_w} } \right)
\end{equation}
where $S_{\mathrm{pen}}$ and $S_{\mathrm{avg}}$ denote penalty and weighted average metrics. NAVSIM develops two evaluation schemes: NAVSIM v1 and NAVSIM v2. In NAVSIM v1, the sub-metrics in $S_{\mathrm{pen}}$ include No Collisions (NC) and Drivable Area Compliance (DAC), while $S_{\mathrm{avg}}$ includes Ego Progress (EP), Time-to-Collision (TTC), and Comfort (C). NAVSIM v2 introduces 4 new sub-metrics: Driving Direction Compliance (DDC) and Traffic Light Compliance (TLC) belong to $S_{\mathrm{pen}}$, while Lane Keeping (LK) and Extended Comfort (EC) belong to $S_{\mathrm{avg}}$.

HUGSIM is a closed-loop planning benchmark covering over 400 simulation scenarios. These scenarios are categorized into four difficulty levels: Easy, Medium, Hard, and Extreme. HUGSIM provides a comprehensive metric, HD-Score, to evaluate the closed-loop performance of planning models across multiple dimensions, including No Collision (NC), Driveable Area Compliance (DAC), Time-to-Collision (TTC), Comfort (COM), and Route Completion ($R_c$).
During the simulation, the HD-Score at each timestep $t$ is calculated as follows:
\begin{equation}
    \text{HD-Score}_t = \left( \prod\limits_{m\in\{NC,DAC\}}{score_m} \right) \cdot \left( \frac{\sum_{w \in \{TTC, COM\}} {weight_w \cdot score_w}}{\sum_{w \in \{TTC, COM\}} weight_w} \right)
\end{equation}

The final HD-Score is obtained by averaging the scores across all timesteps and multiplying the route completion rate $R_c$.

\mypara{Model Details}
The backbone of \modelName is identical to Transfuser~\cite{chitta2022transfuser}. The environment encoder $\mathrm{Enc_{env}}$ adopts a dual-modal pipeline, both the image and LiDAR branches utilize a ResNet34~\cite{he2016deep} backbone. The resulting environment feature $cond$ comprises encoded query vectors of other agents and an $8 \times 8$ BEV feature map. The trajectory encoder $\mathrm{Enc_{traj}}$ is a 2-layer MLP, while both the Driving Intention Decoder $\mathrm{Dec_{gbl}}$ and Trajectory Refinement Decoder $\mathrm{Dec_{lcl}}$ consist of 1 Transformer layer. For the Hierarchical Diffusion Policy, we set the number of predefined candidate trajectories $M_1$ to 20. After the first-stage denoising, $K=2$ top-scoring trajectories are selected. In the trajectory expansion algorithm, the expansion factor $n_{\mathrm{exp}}$ is set to 5, the corresponding radial scaling coefficient set $\{\lambda_u\}$ is $\{0.92, 0.96, 1.0, 1.04, 1.08\}$, and the angular coefficient set $\{\delta_v\}$ is $\{-6^\circ, -3^\circ, 0^\circ, 3^\circ, 6^\circ\}$. This process generates $M_{\mathrm{sub}} = 25$ trajectories within each local region and $M_2 = 50$ trajectories in total. The trajectory vocabulary $\mathcal{V}$ for the Offline Reward Approximation Scheme consists of $N = 8192$ trajectories. The coefficients $\gamma_{\mathrm{dist}}$, $\gamma_{\mathrm{pdms}}$, and $\gamma_{\mathrm{rl}}$ in Equ.~\ref{equ:lcl_score_infer} are 0.6, 0.05, and 0.01. We train two model variants: \modelName and \modelName-L. \modelName requires both image and LiDAR inputs. \modelName-L follows Latent Transfuser~\cite{chitta2022transfuser}, replacing LiDAR inputs with learnable positional embeddings, thus only requiring image inputs. More model details and training details are shown in the Appendix~\ref{sec:appendix-imple_details}.

\begin{table*}[!t]
    \centering
    \small
    \caption{Results on NAVSIM v1. ``C+L'' denotes that the model requires both RGB image and LiDAR as inputs, while ``C'' indicates that the model only receives image input.
    }
    \begin{tabular}{c|c|*{6}{c}}
        \toprule
        Method & Input & NC $\uparrow$ & DAC $\uparrow$ & EP $\uparrow$ & TTC $\uparrow$ & C $\uparrow$ & PDMS $\uparrow$ \\
        \midrule
        Human & — & 100 & 100 & 87.5 & 100 & 99.9 & 94.8 \\
        \midrule
        Transfuser~\cite{chitta2022transfuser} & C+L & 97.7 & 92.8 & 79.2 & 92.8 & 100 & 84.0 \\
        UniAD~\cite{hu2023planning} & C & 97.8 & 91.9 & 78.8 & 92.9 & 100 & 83.4 \\
        VADv2~\cite{chen2024vadv2} & C & 97.9 & 91.7 & 77.6 & 92.9 & 100 & 83.0 \\
        LAW~\cite{li2025enhancing} & C & 96.4 & 95.4 & 81.7 & 88.7 & 99.9 & 84.6 \\
        DRAMA~\cite{yuan2024drama} & C+L & 98.0 & 93.1 & 80.1 & 94.8 & 100 & 85.5 \\
        GoalFlow~\cite{xing2025goalflow} & C+L & 98.3 & 93.8 & 79.8 & 94.3 & 100 & 85.7 \\
        Hydra-MDP~\cite{li2024hydra} & C+L & 98.3 & 96.0 & 78.7 & 94.6 & 100 & 86.5 \\
        DiffusionDrive~\cite{liao2024diffusiondrive} & C+L & 98.2 & 96.2 & 82.2 & 94.7 & 100 & 88.1 \\
        WoTE~\cite{li2025endtoend} & C+L & 98.5 & 96.8 & 81.9 & 94.9 & 99.9 & 88.3 \\
        DriveSuprim~\cite{yao2025drivesuprim} & C & 97.8 & 97.3 & 86.7 & 93.6 & 100 & \underline{89.9} \\
        \midrule
        \textbf{\modelName} & C+L & 98.2 & 97.3 & 87.4 & 97.5 & 100 & \textbf{90.2} \\
        \textbf{HAD-L} & C & 98.1 & 97.2 & 87.2 & 97.3 & 100 & \underline{89.9} \\
        \bottomrule
    \end{tabular}
    \label{tab:navsim_v1}
    \vspace{-5pt}
\end{table*}

\vspace{1em}

\begin{table*}[!t]
    \centering
    \small
    \caption{Results on NAVSIM v2. ``C+L'' denotes that the model requires both RGB image and LiDAR as inputs, while ``C'' indicates that the model only receives image input.}
    \resizebox{\textwidth}{!}{
    \begin{tabular}{c|c|*{9}{c}|cc}
        \toprule
        Method & Input & NC $\uparrow$ & DAC $\uparrow$ & DDC $\uparrow$ & TLC $\uparrow$ & EP $\uparrow$ & TTC $\uparrow$ & LK $\uparrow$ & HC $\uparrow$ & EC $\uparrow$ & EPDMS $\uparrow$ \\
        \midrule
        Human & — & 100 & 100 & 99.8 & 100 & 87.4 & 100 & 100 & 98.1 & 90.1 & 90.3 \\
        \midrule
        Ego Status MLP & — & 93.1 & 77.9 & 92.7 & 99.6 & 86.0 & 91.5 & 89.4 & 98.3 & 85.4 & 64.0 \\
        Transfuser~\cite{chitta2022transfuser} & C+L & 96.9 & 89.9 & 97.8 & 99.7 & 87.1 & 95.4 & 92.7 & 98.3 & 87.2 & 76.7 \\
        HydraMDP++~\cite{li2024hydramdp_pp} & C & 97.2 & 97.5 & 99.4 & 99.6 & 83.1 & 96.5 & 94.4 & 98.2 & 70.9 & 81.4 \\
        DriveSuprim~\cite{yao2025drivesuprim} & C & 97.5 & 96.5 & 99.4 & 99.6 & 88.4 & 96.6 & 95.5 & 98.3 & 77.0 & 83.1 \\
        DiffusionDriveV2~\cite{zou2025diffusiondrivev2} & C+L & 97.7 & 96.6 & 99.2 & 99.8 & 88.9 & 97.2 & 96.0 & 97.8 & 91.0 & 85.5 \\
        DiffRefiner~\cite{yin2025diffrefiner} & C & 98.5 & 97.4 & 99.6 & 99.8 & 87.6 & 97.7 & 97.7 & 98.3 & 86.2 & 86.2 \\
        EvaDrive~\cite{jiao2025evadrive} & C & 98.8 & 98.5 & 98.9 & 99.8 & 96.6 & 98.4 & 94.3 & 97.8 & 55.9 & 86.3 \\
        \midrule
        \textbf{\modelName} & C+L & 98.2 & 97.3 & 99.2 & 99.8 & 87.4 & 97.5 & 95.2 & 98.3 & 86.2 & \textbf{88.6} \\
        \textbf{HAD-L} & C & 98.1 & 97.2 & 99.3 & 99.8 & 87.2 & 97.3 & 95.3 & 98.3 & 86.4 & \underline{88.5} \\
        \bottomrule
    \end{tabular}
    }
    \label{tab:navsim_v2}
    \vspace{-10pt}
\end{table*}


\subsection{Quantitative Results}

\mypara{Result on NAVSIM}
Tab.~\ref{tab:navsim_v1} and Tab.~\ref{tab:navsim_v2} present the performance of \modelName on the NAVSIM benchmark. On NAVSIM v1, \modelName achieves 90.2 PDMS, outperforming DiffusionDrive by 2.1 PDMS. On the more challenging NAVSIM v2 benchmark, \modelName reaches 88.6 EPDMS, surpassing previous state-of-the-art methods such as DiffRefiner by 2.4 and EvaDrive by 2.3 PDMS, respectively. Furthermore, the camera-only variant \modelName-L achieves nearly identical performance to \modelName on NAVSIM v1 and v2, with gaps of only 0.3\% and 0.1\%, demonstrating robustness across different input modalities. The inference speed is reported in Appendix~\ref{subsec:appendix-infer_speed}.

\begin{table*}[!t]
    \centering
    \small
    \caption{Results on the HUGSIM dataset. ``RC'' denotes the Route Completion rate, and ``HDS'' represents the HD-Score metric. The asterisk symbol ``*'' indicates that the results are evaluated on both public and private datasets; otherwise, the results refer to the performance on the public dataset only.
    }
    \begin{tabular}{c|cccccccc|cc}
    \toprule
    \multirow{2}{*}{Method} 
    & \multicolumn{2}{c}{Easy} 
    & \multicolumn{2}{c}{Medium} 
    & \multicolumn{2}{c}{Hard} 
    & \multicolumn{2}{c|}{Extreme} 
    & \multicolumn{2}{c}{Overall} \\
    \cmidrule{2-3} \cmidrule{4-5} \cmidrule{6-7} \cmidrule{8-9} \cmidrule{10-11}
     & RC & HDS & RC & HDS & RC & HDS & RC & HDS & RC & HDS \\
    \midrule
    UniAD*~\cite{hu2023planning} & 58.6 & 48.7 & 41.2 & 29.5 & 40.4 & 27.3 & 26.0 & 14.3 & 40.6 & 28.9 \\
    VAD*~\cite{jiang2023vad} & 38.7 & 24.3 & 27.0 & 9.9 & 25.5 & 10.4 & 23.0  & 8.2 & 27.9 & 12.3 \\
    Latent TransFuser*~\cite{chitta2022transfuser} & 68.4 & 52.8 & 40.7 & 24.6 & 36.9 & 19.8 & 25.5  & 8.1 & 41.4 & 24.8 \\
    \midrule
    Latent TransFuser~\cite{chitta2022transfuser} & 60.4 & 42.5 & 39.4 & 17.7 & 32.7 & 11.8 & 27.9 & 10.6 & 37.9 & 18.0  \\
    GTRS-Dense~\cite{li2025generalized} & 64.2 & 55.5 & 50.0 & 39.0 & 20.7 & 11.7 & 22.3  & 14.3 & 38.0  & 28.6 \\
    ZTRS~\cite{li2025ztrs} & 74.4 & 60.8 & 50.9 & 34.2 & 32.7 & 20.5 & 21.9 & 11.0 & 42.6 & 28.9 \\
    \midrule
    \textbf{HAD-L} & 76.9 & 65.9 & 49.3 & 31.4 & 36.4 & 18.0 & 39.1 & 22.5 & \textbf{47.5} & \textbf{30.8} \\
    \bottomrule
    \end{tabular}
    \label{tab:hugsim}
\end{table*}

\mypara{Result on HUGSIM}
Tab.~\ref{tab:hugsim} presents the evaluation results of \modelName-L on the high-fidelity closed-loop benchmark HUGSIM.
\modelName-L achieves 47.5 Route Completion and 30.8 HD-Score. Compared to ZTRS, \modelName-L improves the RC and HD-Score by 4.9 and 1.9 points. The performance improvement becomes even more pronounced in challenging scenarios. In extreme scenarios characterized by frequent and aggressive agent attacks, \modelName-L maintains an RC of 39.1 and an HD-Score of 22.5, significantly surpassing all baseline methods by a huge margin. These results highlight the model's exceptional driving capabilities in highly interactive, challenging environments.

\subsection{Ablation Studies}

\newcolumntype{C}{>{\centering\arraybackslash}X}

\begin{figure}[t!]
\hspace{1.2em}
\begin{minipage}[b]{0.45\textwidth}
  \centering
  \small
  \setlength{\tabcolsep}{4pt}
  \captionof{table}{Ablation on different trajectory evaluation metrics in hierarchical denoising stages.}
  \label{tab:ablation-traj_score}
  \resizebox{\linewidth}{!}
  {
  \renewcommand{\arraystretch}{1.19}
  \begin{tabular}{ccccccc}
    \toprule
    \multicolumn{3}{c}{Global Denoising} & \multicolumn{3}{c}{Local Denoising} & \multirow{2}{*}{\makecell[c]{EPDMS $\uparrow$}} \\
    \cmidrule(lr){1-3} \cmidrule(lr){4-6}
    Dist. & Safety & RL & Dist. & Safety & RL & \\
    \midrule
    \checkmark &        &        & \checkmark &        &        & 86.7 \\
    \checkmark &        &        & \checkmark & \checkmark &     & 87.3 \\
    \checkmark &        &        & \checkmark & \checkmark & \checkmark & \textbf{88.6} \\
    \checkmark & \checkmark & \checkmark & \checkmark & \checkmark & \checkmark & 87.2 \\
    \bottomrule
  \end{tabular}
  }
\end{minipage}
\hfill
\begin{minipage}[b]{0.4\textwidth}
  \centering
  \small
  \setlength{\tabcolsep}{5pt}
  \captionof{table}{Ablation on the number of first-stage selected trajectories and expanded trajectories.}
  \label{tab:ablation-traj_expansion-topk_num}
  \resizebox{\linewidth}{!}
  {
  \renewcommand{\arraystretch}{1.02}
  \begin{tabular}{ccccc}
    \toprule
    \multicolumn{2}{c}{Training} & \multicolumn{2}{c}{Inference} & \multirow{2}{*}{\makecell[c]{EPDMS $\uparrow$}} \\
    \cmidrule(lr){1-2} \cmidrule(lr){3-4}
    Top-K & $n_{\mathrm{exp}}^2$ & Top-K & $n_{\mathrm{exp}}^2$ & \\
    \midrule
    1  & $5\times5$ & 1  & $5\times5$ & 88.1 \\
    2  & $5\times5$ & 2  & $5\times5$ & 88.5 \\
    2  & $5\times5$ & 2  & $7\times7$ & \textbf{88.6} \\
    4  & $5\times5$ & 4  & $5\times5$ & 86.4 \\
    20 & $5\times5$ & 20 & $5\times5$ & 79.8 \\
    \bottomrule
  \end{tabular}
  }
\end{minipage}
\hspace{1.2em}
\end{figure}

\mypara{Trajectory Metric Scores in Hierarchical Modeling}
In our hierarchical modeling, we use the distance towards the human expert trajectory to select top-K candidates in the Driving Intention Establishment stage, while utilizing multiple metrics in the Local Trajectory Refinement stage, as shown in Equ.~\ref{equ:dec_lcl}.
The result in Tab.~\ref{tab:ablation-traj_score} shows the influence of different metric scores within the hierarchical framework, including the distance, safety metric scores, and RL weights.
When the model only utilizes the distance to the human trajectory as supervision in trajectory refinement, it only achieves an 86.7 EPDMS. Incorporating the safety metric as scoring supervision raises the performance to 87.3 EPDMS. Furthermore, adding reinforcement learning training in the Local Trajectory Refinement stage further improves the EPDMS to 88.6. This indicates that using comprehensive metrics and adopting RL training helps the model learns a more robust driving policy aligning with driving rules.
Moreover, introducing safety metric scoring and RL supervision in the Driving Intention Establishment stage leads to a performance drop to 87.2. This indicates that the whole unstructured driving space makes it difficult for complex reward signals to provide valuable information.

\mypara{Trajectory Expansion Algorithm}
Our proposed Structure-preserved Trajectory Expansion Algorithm effectively explores the local driving region while preserving the trajectory kinematic structure. Tab.~\ref{tab:ablation-traj_expansion-topk_num} and Tab.~\ref{tab:ablation-traj_expansion} further show the superiority.
Tab.~\ref{tab:ablation-traj_expansion-topk_num} shows the ablation on the number of first-stage filtered trajectories $K$ and the expansion scale $n_{\mathrm{exp}} \times n_{\mathrm{exp}}$. The results show that constraining $K$ to a small range (e.g., $K=2$) significantly enhances model performance, achieving the optimal 88.6 EPDMS. When $K$ is increased to include all 20 trajectories, the EPDMS drops to 79.8. This indicates the necessity to adopt hierarchical modeling in end-to-end planning. Furthermore, increasing the sampling density from $5\times5$ to $7\times7$ during inference yields a performance gain of 0.1 EPDMS. This suggests that a denser local area search during inference helps the model precisely locate better trajectories.

\begin{figure}[!t]
    \centering
    \includegraphics[width=0.9\textwidth]{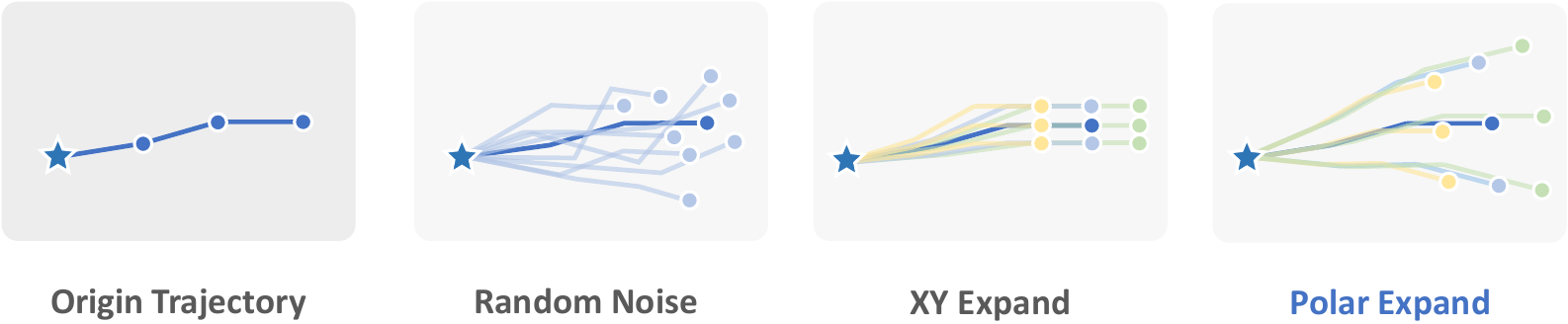}
    \caption{Illustration of different trajectory expansion algorithms. Directly adding random noise is harmful to trajectory kinematic structure, expanding trajectory in the Cartesian space (XY Expand) suffer from insufficient exploration on local region, and expansion in the polar space (Polar Expand) leads to comprehensive exploration.}
    \label{fig:traj_expand_compare}
\end{figure}

Tab.~\ref{tab:ablation-traj_expansion} compares our trajectory expansion algorithm with other approaches illustrated in Fig.~\ref{fig:traj_expand_compare}. Directly adding random noise to trajectories damages their intrinsic structure, only leading to 84.9 EPDMS. Expanding the trajectory in the Cartesian coordinate system~\cite{zou2025diffusiondrivev2} (XY Expand) preserves the trajectory structure, but fails to provide a spatially balanced exploration. As shown in Fig.~\ref{fig:traj_expand_compare}, when the trajectory y-coordinate is small, the exploration range along the y-axis becomes severely restricted. In comparison, our expansion in the polar coordinate system can make a sufficient coverage of the local area, leading to the best 88.6 EPDMS.

\begin{table*}[!t]
    \centering
    \small
    \caption{Comparison of different trajectory expansion algorithms.}
    \label{tab:ablation-traj_expansion}
    \begin{tabular}{c|*{9}{c}|cc}
        \toprule
        Expansion & NC $\uparrow$ & DAC $\uparrow$ & DDC $\uparrow$ & TLC $\uparrow$ & EP $\uparrow$ & TTC $\uparrow$ & LK $\uparrow$ & HC $\uparrow$ & EC $\uparrow$ & EPDMS $\uparrow$ \\
        \midrule
        Random Noise & 98.1 & 93.3 & 99.3 & 99.8 & 87.6 & 96.9 & 95.6 & 98.3 & 85.4 & 84.9 \\
        XY Expand & 98.2 & 93.8 & 99.4 & 99.8 & 87.6 & 97.1 & 95.4 & 98.3 & 87.5 & 85.9 \\
        Polar Expand & 98.2 & 97.3 & 99.2 & 99.8 & 87.4 & 97.5 & 95.2 & 98.3 & 86.2 & \textbf{88.6} \\
        \bottomrule
    \end{tabular}
    \vspace{-10pt}
\end{table*}

\begin{table*}[!t]
    \centering
    \small
    \vspace{1em}
    \caption{Ablation on the effectiveness of MDPO. Results show that Metric-Decoupled Policy Optimization improves overall performance compared to using a coupled reward and optimization.}
    \resizebox{\textwidth}{!}{
    \begin{tabular}{c|c|*{9}{c}|cc}
        \toprule
        Reward & Decoupled & NC $\uparrow$ & DAC $\uparrow$ & DDC $\uparrow$ & TLC $\uparrow$ & EP $\uparrow$ & TTC $\uparrow$ & LK $\uparrow$ & HC $\uparrow$ & EC $\uparrow$ & EPDMS $\uparrow$ \\
        \midrule
        Dist. & & 92.0 & 91.2 & 96.6 & 99.5 & 88.7 & 91.6 & 85.1 & 41.3 & 3.9 & 62.6 \\
        PDMS & & 97.9 & 96.7 & 99.3 & 99.7 & 88.1 & 97.1 & 94.7 & 98.2 & 84.5 & 87.8 \\
        Dist. + PDMS & &  97.9 & 96.2 & 99.1 & 99.8 & 88.0 & 96.9 & 94.3 & 98.3 & 80.8 & 86.9 \\
        \midrule
        Decoupled Metrics & \CM & 98.2 & 97.3 & 99.2 & 99.8 & 87.4 & 97.5 & 95.2 & 98.3 & 86.2 & \textbf{88.6} \\
        \bottomrule
    \end{tabular}
    }
    \label{tab:ablation-rl_reward}
    \vspace{-10pt}
\end{table*}

\mypara{Superiority of MDPO}
Tab.~\ref{tab:ablation-rl_reward} investigates the RL setting, showing the superiority of MDPO. Firstly, we validate the reward choice when adopting a single RL weight head. The first three rows in Tab.~\ref{tab:ablation-rl_reward} reveal that adopting the comprehensive PDMS score as the reward function can help the model achieve the optimal 87.8 EPDMS. After adopting our proposed MDPO to decouple different metric rewards, the performance is further improved to 88.6 EPDMS, which shows that reward decoupling leads to a better driving policy.


\subsection{Training Efficiency Analysis}

According to our testing, the training efficiency of the model is significantly enhanced by the proposed Offline RL Reward Retrieval Scheme. By utilizing this retrieval mechanism, the reward acquisition latency for a single trajectory is reduced from 0.2449s to 0.0042s. Consequently, 
the total training duration is decreased from 64.4 hours to 13.6 hours. This shows that, compared with online simulation-based training such as DiffusionDriveV2~\cite{zou2025diffusiondrivev2}, our scheme can achieve a 5$\times$ speedup, demonstrating superior efficiency.

\section{Conclusion}

In this paper, we propose \modelName to address the critical limitations of current end-to-end planners on the optimization burden in large action spaces and the challenges of online reinforcement learning. \modelName introduces a Hierarchical Diffusion Policy to decompose planning into coarse-to-fine denoising stages, a Structure-preserved Trajectory Expansion algorithm to generate diverse trajectory candidates while keeping their kinematic structure, and a Metric-Decoupled Policy Optimization (MDPO) framework with an Offline Reward Retrieval Scheme to efficiently learn a fine-grained driving policy.
Experimental results on the NAVSIM and HUGSIM benchmarks demonstrate the superior performance of \modelName on both open-loop and closed-loop planning.

\newpage
\appendix
\section*{Appendix}

\section{Detailed Process of Structure-Preserved Trajectory Expansion}
\label{sec:appendix-traj_expansion}

In this section, we provide the detailed algorithm process of the Structure-Preserved Trajectory Expansion Algorithm in Sec.~\ref{subsec:hierarchical_diffusion_strategy}.

Given a trajectory $\tau=\{(x_t, y_t)\}_{t=1}^T$ and an expansion number $n_\mathrm{exp}$, the Structure-Preserved Trajectory Expansion Algorithm outputs an expanded trajectory set $\mathcal{T}_{\mathrm{exp}} = \{ \tau_{j}\}_{j=1}^{n_{\mathrm{exp}} \cdot n_{\mathrm{exp}}}$. Specifically, the algorithm first projects the Cartesian coordinates of $\tau$ into the polar domain $\tau_{\mathrm{polar}}=\{(\rho_t, \theta_t)\}_{t=1}^T$, then it defines a set of radial scaling coefficients $\{\lambda_u\}_{u=1}^{n_\mathrm{exp}}$ and angular offset coefficients $\{\delta_v\}_{v=1}^{n_\mathrm{exp}}$. The algorithm traverses these coefficients to transform the original trajectory and get multiple expanded trajectories. The detailed algorithm process is shown in Alg.~\ref{alg:traj_expansion}.

\section{Loss Function Details}
\label{sec:appendix-loss_func}

This section provides details of the loss function $L$ used for model training. The loss of \modelName comprises three components: the perception loss $L_{\mathrm{percept}}$, the loss for the Driving Intention Establishment stage $L_{\mathrm{global}}$, and the loss for the Local Trajectory Refinement stage $L_{\mathrm{local}}$.

\mypara{Perception Loss}
The perception loss $L_{\mathrm{percept}}$ follows the Transfuser~\cite{chitta2022transfuser} design and incorporates several auxiliary perception tasks, including a detection loss $L_{\mathrm{box}}$ for 3D bounding box regression, a classification loss $L_{\mathrm{label}}$ for predicting the category labels of other agents, and a BEV semantic segmentation loss $L_{\mathrm{bev}}$:
\begin{equation}
\label{equ:loss_percept}
L_{\mathrm{percept}} = \lambda_{\mathrm{box}} L_{\mathrm{box}} + \lambda_{\mathrm{label}} L_{\mathrm{label}} + \lambda_{\mathrm{bev}} L_{\mathrm{bev}}
\end{equation}
specifically, $L_{\mathrm{box}}$ employs an $L_1$ loss to optimize the location of the bounding boxes of other agents, $L_{\mathrm{label}}$ employs cross-entropy loss to distinguish different categories such as vehicles and pedestrians, and $L_{\mathrm{bev}}$ improves the model’s understanding of drivable areas through pixel-level semantic mask prediction.

\mypara{Driving Intention Establishment Loss}
The loss for the Driving Intention Establishment stage $L_{\mathrm{global}}$ is used to supervise the output of the first-stage decoder $\mathrm{Dec_{gbl}}$, following an anchor-based diffusion planner~\cite{liao2024diffusiondrive}. For the $M_1$ initially generated trajectories, the loss function consists of a trajectory regression loss $L_{\mathrm{gbl, reg}}$ and a region classification loss $L_{\mathrm{gbl, cls}}$:
\begin{gather}
    L_{\mathrm{gbl,reg}} = \sum_{j=1}^{M_1} \mathbf{1}\left(j \in \mathcal{I}_{\mathrm{pos}}\right)
    \left\lVert \hat{\tau}_{\mathrm{gbl},j} - \tau_{\mathrm{gt}} \right\rVert_2^2 \\
    L_{\mathrm{gbl,cls}} = \mathrm{CrossEntropy}
    \left(\hat{\mathbf{s}}_{\mathrm{gbl}}^{(c)},\, \mathbf{y}\right) \\
    \label{equ:loss_global}
    L_{\mathrm{global}} = \lambda_{\mathrm{reg}} L_{\mathrm{gbl,reg}} + \lambda_{\mathrm{cls}} L_{\mathrm{gbl,cls}}
\end{gather}
where $\tau_{\mathrm{gt}}$ represents the human expert trajectory, $\hat{\tau}_{\mathrm{gbl},j}$ is the $j$-th denoised candidate trajectory, and $\mathcal{I}_{\mathrm{pos}}$ denotes the index of the denoised trajectory that is located in the same driving sub-region as the expert trajectory. $\hat{\mathbf{s}}_{\mathrm{gbl}}^{(c)}=\left(\hat{s}_{\mathrm{gbl,1}}^{(c)}, \hat{s}_{\mathrm{gbl,2}}^{(c)}, \dots, \hat{s}_{\mathrm{gbl,M_1}}^{(c)}\right)$ represents the predicted classification scores for each trajectory, and $\mathbf{y}=\left(y_1, y_2, \dots, y_{M_1}\right)$ is the ground truth label for region classification, where each element $y_j$ indicates whether the $j$-th candidate trajectory lies in the same driving sub-region as the expert trajectory.

\begin{algorithm}[!t]
    \caption{Structure-Preserved Trajectory Expansion}
    \label{alg:traj_expansion}
    \KwIn{Trajectory $\tau = \{(x_t, y_t)\}_{t=1}^T$, expansion number $n_{\mathrm{exp}}$, radial scaling coefficient set $\{\lambda_u\}_{u=1}^{n_{\mathrm{exp}}}$, angular offset coefficient set $\{\delta_v\}_{v=1}^{n_{\mathrm{exp}}}$}
    \KwOut{Expanded trajectory set $\mathcal{T}_{\mathrm{exp}} = \{ \tau_{j}\}_{j=1}^{n_{\mathrm{exp}} \cdot n_{\mathrm{exp}}} = \{ \tau^{(u,v)}\}_{u=1, v=1}^{n_{\mathrm{exp}}, n_{\mathrm{exp}}}$}
    
    \tcp{Convert Cartesian coordinates to polar coordinates}
    $\tau_{\mathrm{polar}} \leftarrow \emptyset$\\
    \For{$t \leftarrow 1$ \KwTo $T$}{
        $\rho_t \leftarrow \sqrt{x_t^2 + y_t^2}$\\
        $\theta_t \leftarrow \operatorname{atan2}(y_t, x_t)$\\
        $\tau_{\mathrm{polar}} \leftarrow \tau_{\mathrm{polar}} \cup \{(\rho_t, \theta_t)\}$\\
    }
    
    \tcp{Perform pattern expansion based on coefficient sets}
    $\mathcal{T}_{\mathrm{exp}} \leftarrow \emptyset$\\
    \For{$u \leftarrow 1$ \KwTo $n_{\mathrm{exp}}$}{
        \For{$v \leftarrow 1$ \KwTo $n_{\mathrm{exp}}$}{
            $\tau^{(u,v)} \leftarrow \emptyset$\\
            \For{$t \leftarrow 1$ \KwTo $T$}{
                $\rho_t^{(u)} \leftarrow \rho_t \cdot \lambda_u$\\
                $\theta_t^{(v)} \leftarrow \theta_t + \delta_v$\\
                
                $x_t^{(u,v)} \leftarrow \rho_t^{(u)} \cdot \cos\theta_t^{(v)}$\\
                $y_t^{(u,v)} \leftarrow \rho_t^{(u)} \cdot \sin\theta_t^{(v)}$\\
                $\tau^{(u,v)} \leftarrow \tau^{(u,v)} \cup \{(x_t^{(u,v)}, y_t^{(u,v)})\}$\\
            }
            $\mathcal{T}_{\mathrm{exp}} \leftarrow \mathcal{T}_{\mathrm{exp}} \cup \{\tau^{(u,v)}\}$\\
        }
    }
    \Return $\mathcal{T}_{\mathrm{exp}}$\\
\end{algorithm}

\mypara{Local Trajectory Refinement Loss}
The loss of the Local Trajectory Refinement stage $L_{\mathrm{local}}$ supervises the refined trajectories output in the second stage with multiple quality metrics and RL weights. In Equ.~\ref{equ:dec_lcl}, $\mathrm{Dec_{lcl}}$ predicts several metrics, including distance, safety metric scores, and metric-decoupled reinforcement learning weights. For the distance score $\hat{s}_{\mathrm{lcl}, j}^{(\mathrm{dist})}$, the model calculates the distance $d_j$ between the predicted trajectory $\hat{\tau}_{\mathrm{lcl},j}$ and the expert trajectory $\tau_{\mathrm{gt}}$, and convert the distance to a soft label $d_{\mathrm{norm}, j}$ using Gaussian kernel and max-normalization. The binary cross-entropy (BCE) loss $L_{\mathrm{dist}}$ is computed between the soft label $d_{\mathrm{norm}, j}$ and the predicted score $\hat{s}_{\mathrm{lcl}, j}^{(\mathrm{dist})}$. Secondly, for the safety metric scores $\hat{s}_{\mathrm{lcl}, j}^{(m)}$, the model uses the reference trajectory scores $s_{\mathrm{ref}, j^*}^{(m)}$ retrieved via the Trajectory Reward Retrieval Scheme (detailed in Sec.~\ref{subsec:trajectory_reward_retrieval}) as the ground truth to calculate the BCE loss $L_{\mathrm{safe}}$. Finally, a reinforcement learning loss $L_{\mathrm{rl}}$ is introduced to maximize the expected reward $J$. The Local Trajectory Refinement loss is shown as follows:
\begin{gather}
    \label{equ:soft_dist_label}
    d_{\mathrm{norm}, j} = \frac{\exp(- \beta \|\hat{\tau}_{\mathrm{lcl},j} - \tau_{\mathrm{gt}}\|_2^2) }{ \max\limits_{i\in \{1,\dots,M_2\}}\{\exp(- \beta \|\hat{\tau}_{\mathrm{lcl},i} - \tau_{\mathrm{gt}}\|_2^2)\}}\\
    L_{\mathrm{dist}} = \sum_{j=1}^{M_2} \mathrm{BCE}( \hat{s}_{\mathrm{lcl}, j}^{(\mathrm{dist})} , d_{\mathrm{norm}, j} )\\
    L_{\mathrm{safe}} = \sum_{m} \sum_{j=1}^{M_2} \mathrm{BCE}(\hat{s}_{\mathrm{lcl}, j}^{(m)}, s_{\mathrm{ref}, j^*}^{(m)}) \\
    L_{\mathrm{rl}} = -J = - \frac{1}{K} \sum_{k=1}^K \sum_m \alpha^{(m)} \sum_{j \in \mathcal{I}_k} p_{j}^{(m)} \cdot \bar{r}_{j}^{(m)} \\
    \label{equ:loss_local}
    L_{\mathrm{local}} = \lambda_{\mathrm{dist}} L_{\mathrm{dist}} + \lambda_{\mathrm{safe}} L_{\mathrm{safe}} + \lambda_{\mathrm{rl}} L_{\mathrm{rl}}
\end{gather}
where $\beta$ denotes the coefficient of the Gaussian function, $M_2$ is the number of expanded trajectories in Local Trajectory Refinement stage, $m$ represents safety evaluation metrics defined by NAVSIM~\cite{dauner2024navsim, cao2025pseudo}, including NC, DAC, DDC, TLC, EP, TTC, LK, HC, $\alpha^{(m)}$ denotes the coefficient used to ensemble RL rewards on decoupled metrics, $\mathcal{I}_k$ denotes trajectory indices in the k-th expansion set, $p^{(m)}$ and $\bar{r}_{j}^{(m)}$ denotes the selection probability and normalized reward, identical to Equ.~\ref{equ:reward_j_k} in the main text.

The overall loss of \modelName is the linear combination of three components mentioned above:
\begin{equation}
\label{equ:loss_all}
L = \lambda_{\mathrm{percept}}L_{\mathrm{percept}} + \lambda_{\mathrm{global}}L_{\mathrm{global}} + \lambda_{\mathrm{local}}L_{\mathrm{local}}
\end{equation}

\section{Model Details and Training Details}
\label{sec:appendix-imple_details}

This section supplies some model details and training details not mentioned in Sec~\ref{subsec:imple_details}.

\mypara{Supplementary Model Details}
In Equ.~\ref{equ:pdms_score_infer} of the main text, \modelName utilizes the following equation to ensemble different predicted metric scores:
\begin{equation}
    \hat{s}_{\mathrm{lcl}, j}^{(\mathrm{pdms})} = \sum_{\mathrm{mp}}{\lambda^{(\mathrm{mp})}\log {\hat{s}_{\mathrm{lcl}, j}^{(\mathrm{mp})}}} + \lambda_{\mathrm{avg}}\log\left({\sum_{\mathrm{ma}}{\lambda^{(\mathrm{ma})}s_{\mathrm{lcl}, j}^{(\mathrm{ma})}}}\right)
\end{equation}
where $mp$ and $ma$ are penalty metrics (NC, DAC, DDC, TLC) and average metrics (TTC, EP, LK, HC). The coefficient $\lambda_{\mathrm{avg}}$ is set to 6.0. The value of other coefficients $\lambda^{(\mathrm{mp})}$ and $\lambda^{(\mathrm{ma})}$ is shown in Tab.~\ref{tab:coefficient-v2}.

\begin{table}[!t]
    \centering
    \small
    \caption{The inference coefficients on each metric of NAVSIM. ``Mul'' denotes the multiplied penalties, and ``Avg'' denotes the weighted averages.}
    \label{tab:coefficient-v2}
    \begin{tabularx}{0.7\textwidth}{
        >{\hsize=1.6\hsize}l 
        *{8}{>{\hsize=0.5\hsize\centering\arraybackslash}X} 
    }
        \toprule
        & \multicolumn{4}{c}{Mul} & \multicolumn{4}{c}{Avg} \\
        \cmidrule(lr){2-5} \cmidrule(lr){6-9}
        & NC & DAC & DDC & TLC & EP & TTC & LK & HC \\
        \midrule
        Coefficient & 0.5 & 0.5 & 0.3 & 0.1 & 5.0 & 5.0 & 2.0 & 1.0 \\
        \bottomrule
    \end{tabularx}
\end{table}

\mypara{Training Details}
\modelName is trained on 8 NVIDIA A100 GPUs for 85 epochs. We employ the AdamW optimizer with an initial learning rate of $6\times10^{-4}$. The learning rate follows a Cosine Annealing schedule with Warmup steps (WarmupCosLR). During the first 3 epochs, the learning rate increases linearly from 0 to $6\times10^{-4}$, then it decays following a cosine curve to $1\times10^{-6}$ in the remaining epochs. The coefficients for the model loss functions are listed in Tab.~\ref{tab:loss_coefficient}.

\begin{table}[t]
    \centering
    \small
    \caption{Coefficients in different loss functions.}
    \label{tab:loss_coefficient}
    \begin{tabular}{ccc}
    \toprule
    Equ. No. & Coefficient & Value \\
    \midrule
    \ref{equ:loss_percept} & $\lambda_{\mathrm{box}},\; \lambda_{\mathrm{label}},\; \lambda_{\mathrm{bev}}$ & 1.0,\; 10.0,\; 14.0 \\
    \ref{equ:loss_global} & $\lambda_{\mathrm{reg}},\; \lambda_{\mathrm{cls}}$ & 8.0,\; 10.0 \\
    \ref{equ:loss_local} & $\lambda_{\mathrm{dist}},\; \lambda_{\mathrm{safe}},\; \lambda_{\mathrm{rl}}$ & 10.0,\; 1.0,\; 1.0 \\
    \ref{equ:loss_all} & $\quad\lambda_{\mathrm{percept}},\; \lambda_{\mathrm{global}},\; \lambda_{\mathrm{local}}\quad$ & 1.0,\; 12.0,\; 12.0 \\
    \bottomrule
    \end{tabular}
\end{table}

\section{Supplementary Experimental Results}

\subsection{Parameters and Inference Speed}
\label{subsec:appendix-infer_speed}
We compare the number of parameters and inference speed of different approaches. The result is shown in Tab.~\ref{tab:inference_speed}. \modelName achieves an inference speed of about 30 FPS on a single NVIDIA A100 GPU with 63M parameters, making it capable of real-time inference.

\begin{table}[!t]
    \centering
    \small
    \caption{Comparison of parameter number and inference speed. ``*'' indicates the EPDMS result is from our re-implementation.}
    \begin{tabular}{c|c|c|cc}
        \toprule
        Model & Input & EPDMS $\uparrow$ & Param. $\downarrow$ & FPS $\uparrow$ \\
        \midrule
        Transfuser~\cite{chitta2022transfuser} & C+L & 76.7 & 56M & 56.4 \\
        HydraMDP++~\cite{li2024hydramdp_pp} & C & 81.4 & 53M & 32.0 \\
        DriveSuprim~\cite{yao2025drivesuprim} & C & 83.1 & 61M & 27.2 \\
        DiffusionDrive~\cite{liao2024diffusiondrive} & C+L & 87.5* & 61M & 42.1 \\
        \midrule
        \textbf{\modelName} & C+L & 88.6 & 63M & 30.4 \\
        \textbf{\modelName-L} & C & 88.5 & 63M & 29.4 \\
        \bottomrule
    \end{tabular}
    \label{tab:inference_speed}
\end{table}

\subsection{Performance on NavHard}

Tab.~\ref{table:navhard} presents the performance of various methods on the NavHard benchmark~\cite{cao2025pseudo}.
NavHard is a highly challenging open-loop evaluation benchmark for end-to-end autonomous driving built upon NAVSIM. It adopts a two-stage evaluation protocol: the first stage consists of standard NAVSIM open-loop testing in real-world scenarios, while the second stage utilizes 3D Gaussian Splatting to synthesize diverse difficult scenarios, such as abnormal driving behaviors, to test model performance.

In Tab.~\ref{table:navhard}, \modelName-L achieves 32.3 EPDMS on the NavHard benchmark, surpassing DiffusionDrive~\cite{yao2025drivesuprim} and Latent Transfuser~\cite{chitta2022transfuser}, which also utilizes Transfuser dual-branch backbone.
However, a significant gap remains when compared to methods such as DriveSuprim~\cite{yao2025drivesuprim} and GTRS-Dense~\cite{li2025generalized}.

\begin{figure*}[!t]
    \centering
    \includegraphics[width=0.9\textwidth]{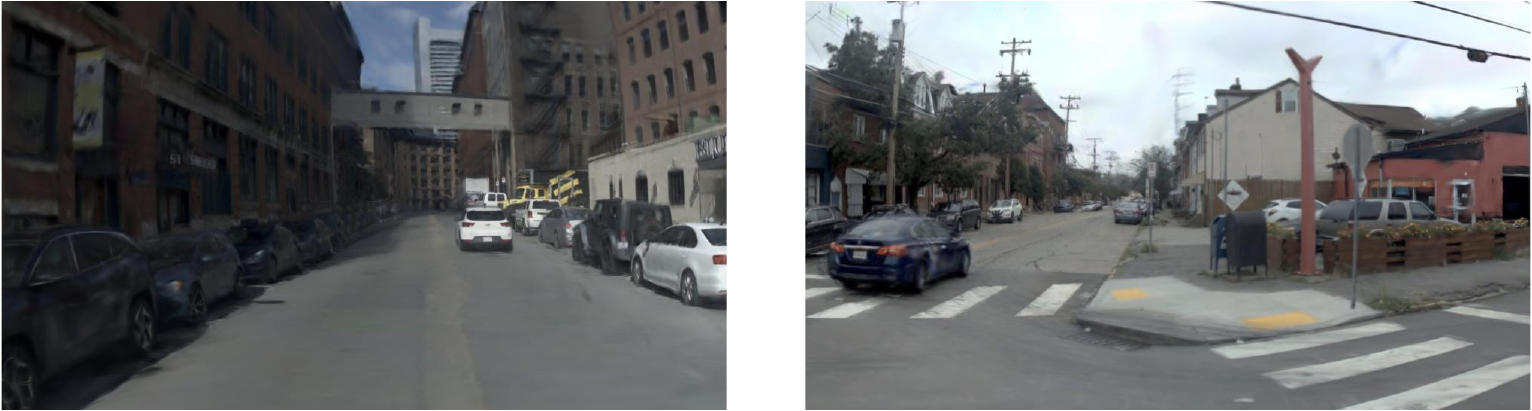}
    \caption{Synthesized data from the second stage of the NavHard benchmark.}
    \label{fig:navhard_data}
\end{figure*}

The reason for this performance gap lies in the high sensitivity of the backbone to sensor noise. \modelName-L adopts the Transfuser-based dual-branch backbone that produces BEV feature maps by fusing image and LiDAR features. As illustrated in Fig.~\ref{fig:navhard_data}, the images synthesized by the Gaussian Splatting model in the second stage of NavHard are imperfect and contain noticeable visual noise, which makes it difficult for the Transfuser backbone to extract representative BEV features. This finding provides a valuable insight for future research. Future works may explore end-to-end planning architectures that do not explicitly rely on BEV feature maps to enhance robustness when facing extreme weather or high-noise sensor inputs.

\begin{table*}[!t]
\small
\centering
\caption{Result on NavHard. ``*'' indicates that the model uses a dual-branch backbone similar to Transfuser.}
\resizebox{\textwidth}{!}{
\begin{tabular}{c|c |c| c c c c c c c c c |c}
    \toprule
    Method 
    & Backbone
    & Stage
    & $\text{NC}$
    & $\text{DAC}$
    & $\text{DDC}$
    & $\text{TLC}$
    & $\text{EP}$
    & $\text{TTC}$
    & $\text{LK}$ 
    & $\text{HC}$ 
    & $\text{EC}$ 
    & $\text{EPDMS}$  \\
    \midrule
    
    PDM-Closed~\cite{dauner2023parting}  & - &   \makecell{Stage 1 \\ Stage 2} & \makecell{94.4 \\ 88.1} & \makecell{98.8 \\ 90.6} & \makecell{100 \\ 96.3} & \makecell{99.5 \\ 98.5} & \makecell{100 \\ 100} & \makecell{93.5 \\ 83.1} & \makecell{99.3 \\ 73.7} & \makecell{87.7 \\ 91.5} & \makecell{36.0 \\ 25.4} & 51.3    \\
    \midrule
    
    LTF~\cite{chitta2022transfuser}  & ResNet34* &   \makecell{Stage 1 \\ Stage 2} &
    \makecell{96.2 \\ 77.7} & \makecell{79.5 \\ 70.2} & \makecell{99.1 \\ 84.2} & \makecell{99.5 \\ 98.0} & \makecell{{84.1} \\ {85.1}} & \makecell{95.1 \\ 75.6} & \makecell{94.2 \\ 45.4} & \makecell{97.5 \\ 95.7} & \makecell{{79.1} \\ {75.9}} & 23.1    \\
    \midrule
    
    \multirow{1}{*}{GTRS-Dense~\cite{li2025generalized}} & V2-99&   \makecell{Stage 1 \\ Stage 2}  &
    \makecell{98.7 \\ 91.4} & \makecell{95.8 \\ 89.2} & \makecell{99.4 \\ 94.4} & \makecell{99.3 \\ 98.8} & \makecell{72.8 \\ 69.5} & \makecell{98.7 \\ 90.1} & \makecell{95.1 \\ 54.6} & \makecell{96.9 \\ 94.1} & \makecell{40.4 \\ 49.7} & 41.7    \\
    
    \midrule

    \multirow{3}{*}{DriveSuprim~\cite{yao2025drivesuprim}}  & ResNet34 &   \makecell{Stage 1 \\ Stage 2}  &
    \makecell{97.2 \\ 88.6} & \makecell{96.2 \\ 86.0} & \makecell{99.3 \\ 89.4} & \makecell{99.6 \\ 98.4} & \makecell{71.4 \\ 74.7} & \makecell{96.7 \\ 86.5} & \makecell{94.7 \\ 55.3} & \makecell{{97.3} \\ 97.7} & \makecell{47.1 \\ 59.4} & 39.5    \\
    \cmidrule{2-13}
      & V2-99&   \makecell{Stage 1 \\ Stage 2}  &
    \makecell{{98.9} \\ 87.9} & \makecell{95.1 \\ 88.8} & \makecell{99.2 \\ 89.6} & \makecell{99.6 \\ 98.8} & \makecell{76.1 \\ 80.3} & \makecell{99.1 \\ 86.0} & \makecell{94.7 \\ 53.5} & \makecell{{97.6} \\ 97.1} & \makecell{54.2 \\ 56.1} & 42.1    \\
    
    \midrule

    DiffusionDrive~\cite{liao2024diffusiondrive}  & ResNet34* &   \makecell{Stage 1 \\ Stage 2} & \makecell{96.8 \\ 80.1} & \makecell{86.0 \\ 72.8} & \makecell{98.8 \\ 84.4} & \makecell{99.3 \\ 98.4} & \makecell{84.0 \\ 85.9} & \makecell{95.8 \\ 76.6} & \makecell{96.7 \\ 46.4} & \makecell{97.6 \\ 96.3} & \makecell{66.7 \\ 40.5} & 27.5  \\
    \midrule

    \textbf{\modelName-L}  & ResNet34* &   \makecell{Stage 1 \\ Stage 2} & \makecell{96.2 \\ 80.2} & \makecell{90.4 \\ 75.2} & \makecell{98.6 \\ 83.7} & \makecell{99.3 \\ 98.4} & \makecell{83.6 \\ 86.6} & \makecell{95.8 \\ 77.0} & \makecell{96.0 \\ 47.2} & \makecell{97.8 \\ 95.9} & \makecell{78.2 \\ 70.5} & 32.3    \\
\bottomrule
\end{tabular}}
\label{table:navhard}
\end{table*}

\begin{figure*}[!t]
    \centering
    \includegraphics[width=0.95\textwidth]{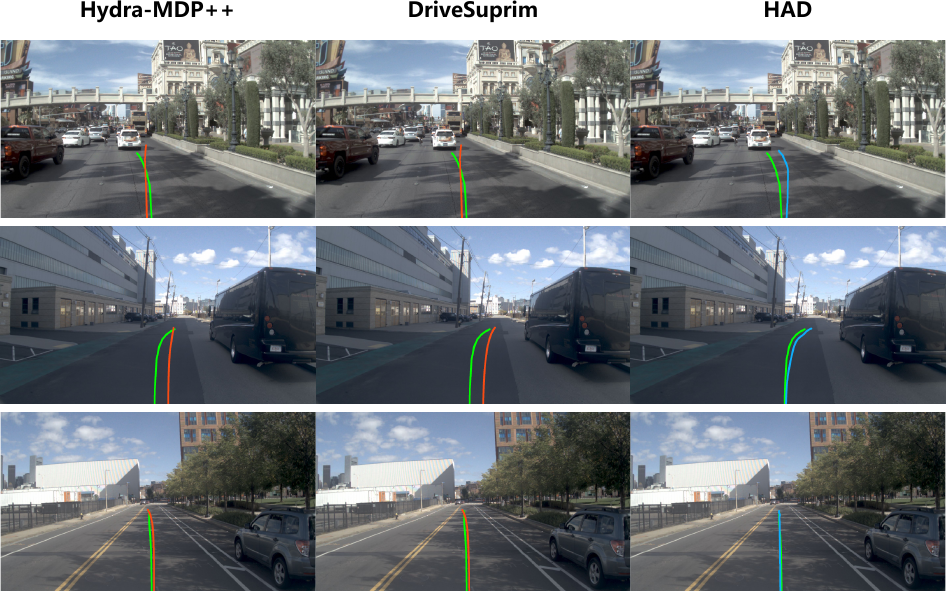}
    \caption{Visualization result on the NAVSIM dataset. In each example, the green trajectory represents the ground truth from the human expert, the red trajectories are generated by other methods, and the blue trajectory is produced by \modelName. Zoom in for a better view.}
    \label{fig:vis_navsim}
\end{figure*}

\subsection{Visualization Results}

\begin{figure*}[!t]
    \centering
    \includegraphics[width=0.86\textwidth]{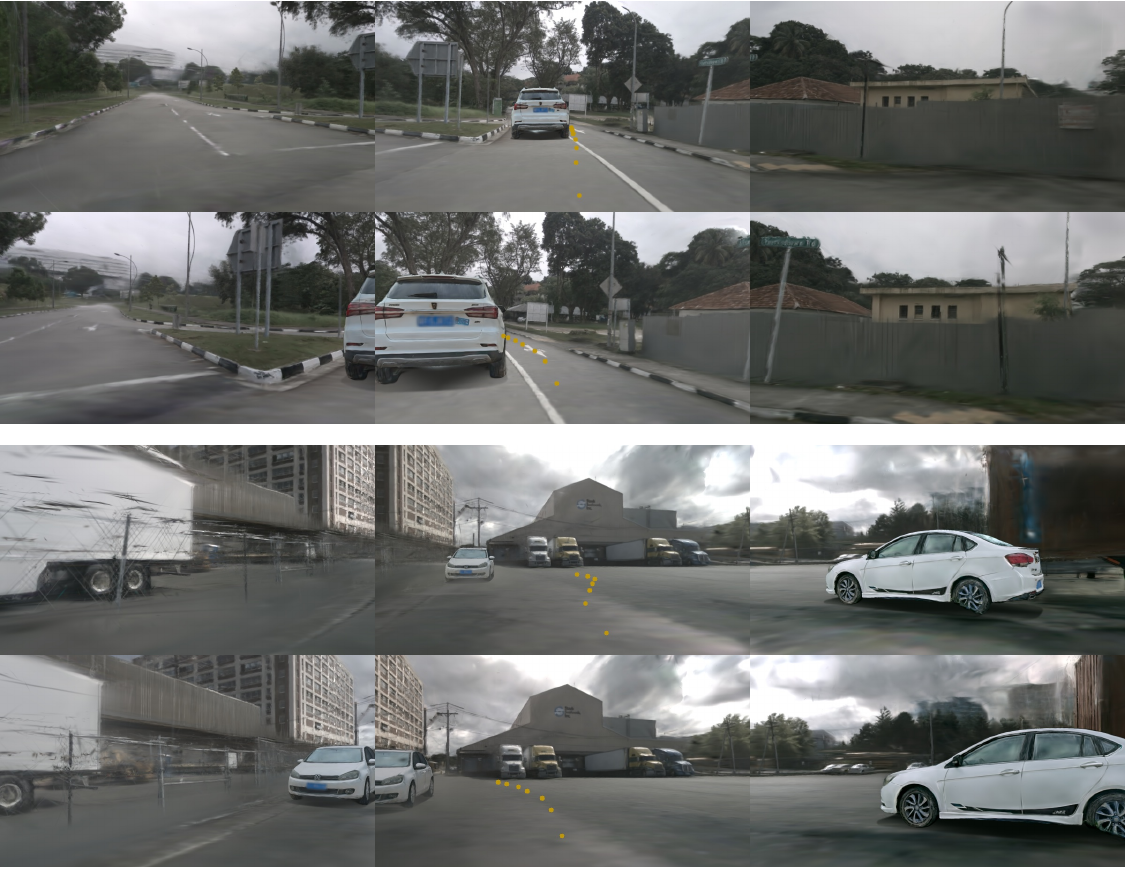}
    \caption{Visualization result on the HUGSIM dataset.}
    \label{fig:vis_hugsim}
\end{figure*}

We provide the visualization results of \modelName in Fig.~\ref{fig:vis_navsim} and Fig.~\ref{fig:vis_hugsim}.

\mypara{Results on NAVSIM}
Fig.~\ref{fig:vis_navsim} illustrates a qualitative comparison between Hydra-MDP++~\cite{li2024hydramdp_pp}, DriveSuprim~\cite{yao2025drivesuprim}, and \modelName across various challenging scenarios. We validate the superiority of our proposed method in complex scene interaction and long-distance driving robustness.

In the complex interaction scenarios shown in the first two rows of Fig.~\ref{fig:vis_navsim}, the ego vehicle must generate precise trajectories to avoid potential collisions. The red trajectories generated by selection-based methods exhibit clear decision-making limitations, resulting in insufficient safety margins from other vehicles. In contrast, \modelName demonstrates superior environmental understanding capability. The blue trajectories generated by our method maintain appropriate safety distances from other vehicles and closely align with the expert trajectories. This indicates that the Hierarchical Diffusion Policy can effectively identify the optimal trajectory within the local driving region.

In the long-distance straight-line scenario shown in the third row, \modelName demonstrates excellent driving direction maintenance. The blue trajectory remains near the lane centerline even at long range, without accumulating decision-making bias over time. It validates that by applying the MDPO algorithm, the model learns a policy with greater long-term robustness than pure supervised learning, effectively preventing horizontal drift in long-distance planning.

\mypara{Results on HUGSIM}
Fig.~\ref{fig:vis_hugsim} visualizes the output of \modelName in the closed-loop HUGSIM benchmark, highlighting the model's real-time avoidance capabilities during dynamic interactive maneuvers.

In the scenarios shown in the first two rows of Fig.~\ref{fig:vis_hugsim}, when the ego vehicle is driving in a single lane and encounters a slow-moving vehicle ahead, \modelName accurately identifies the available space on the right and plans an avoidance trajectory that shifts toward that side. In the more complex intersection scenarios presented in the bottom two rows, \modelName demonstrates strong decision-making capability when facing oncoming traffic. The model adaptively adjusts its speed and direction, generating a smooth bypassing trajectory.

The robust driving behavior observed in closed-loop planning further validates the effectiveness of the Metric-Decoupled Policy Optimization algorithm. The reinforcement learning strategy enables the model to make decisions that comply with multiple driving rules, thereby ensuring safe and smooth vehicle behavior in highly dynamic environments.

\clearpage

\bibliographystyle{plainnat}
\bibliography{main}






\end{document}